%% file: paper.tex
\documentclass[runningheads]{llncs}

\usepackage[T1]{fontenc}

\usepackage{color,graphicx,amsmath,amssymb,float,multirow,booktabs,hyperref,colortbl,subfigure,cite,enumitem,multicol}

\hypersetup{colorlinks=true}

\begin{document}

\title{Cross-Domain Local Characteristic Enhanced Deepfake Video Detection}

\author{
Zihan Liu \and
Hanyi Wang \and
Shilin Wang\thanks{Corresponding Author}}

\titlerunning{Cross-Domain Local Forensics}
\authorrunning{Z. Liu et al.}

\institute{
School of Electronic Information and Electrical Engineering,\\
Shanghai Jiao Tong University\\
\email{\{lzh123,why\_820,wsl\}@sjtu.edu.cn}}

\maketitle              

\input{contents/abstract}
\input{contents/introduction}
\input{contents/related-work}
\input{contents/method}
\input{contents/experiment}
\input{contents/conclusion}

\bibliographystyle{splncs04}
\bibliography{egbib}

\end{document}

%% file: contents/abstract.tex
\begin{abstract}

As ultra-realistic face forgery techniques emerge, deepfake detection has attracted increasing attention due to security concerns. Many detectors cannot achieve accurate results when detecting unseen manipulations despite excellent performance on known forgeries. In this paper, we are motivated by the observation that the discrepancies between real and fake videos are extremely subtle and localized, and inconsistencies or irregularities can exist in some critical facial regions across various information domains. To this end, we propose a novel pipeline, Cross-Domain Local Forensics (XDLF), for more general deepfake video detection. In the proposed pipeline, a specialized framework is presented to simultaneously exploit local forgery patterns from space, frequency, and time domains, thus learning cross-domain features to detect forgeries. Moreover, the framework leverages four high-level forgery-sensitive local regions of a human face to guide the model to enhance subtle artifacts and localize potential anomalies. Extensive experiments on several benchmark datasets demonstrate the impressive performance of our method, and we achieve superiority over several state-of-the-art methods on cross-dataset generalization. We also examined the factors that contribute to its performance through ablations, which suggests that exploiting cross-domain local characteristics is a noteworthy direction for developing more general deepfake detectors.

\end{abstract}

%% file: contents/introduction.tex
\section{Introduction}

Recent years have witnessed tremendous progress in face forgery techniques~\cite{19fsgan,20faceshifter,20cdf,20dfdc}, i.e., deepfake, due to the emergence of deep generative models. As such techniques can synthesize highly realistic fake videos without considerable human effort, they can easily be abused by malicious attackers to counterfeit imperceptible identities or behaviors, thereby causing severe political and social threats. To mitigate such threats, numerous automatic deepfake detection methods~\cite{21lipforensics,21multiattention,21hff,21spsl,21localrelation,21pcl,21idreveal,21dcl} have been proposed.

Most studies formulated deepfake detection as a binary classification problem with global supervision (i.e., real/fake) for training. They relied on convolutional neural networks (CNN) to extract discriminative features to detect forgeries. While these methods achieved satisfactory accuracy when the training and test sets have similar distributions, their performance significantly dropped when encountering novel manipulations. Therefore, many works~\cite{20twobranch,20facexray,21lipforensics,21hff} aimed at improving generalization to unseen forgeries with diverse approaches.

With the continuous refinement of face forgery methods, the discrepancies between real and fake videos are increasingly subtle and localized. Inconsistencies or irregularities can exist in some critical local regions across various information domains, e.g., space~\cite{18mesonet,19ff}, frequency~\cite{20f3net,21localrelation,21spsl,21hff}, and time~\cite{19cnngru,19opticalflow,21dianet,21inconsistency} domains. However, these anomalies are so fine-grained that vanilla CNN often fails to capture them. Many detection algorithms exploited local characteristics to enhance generalization performance. However, these algorithms still had some limitations in representing local features. On the one hand, some algorithms~\cite{21lipforensics} solely relied on a specific facial region to distinguish between real and fake videos while ignoring other facial regions, which restricted the detection performance. On the other hand, many algorithms~\cite{20twobranch,21hff,21localrelation} made insufficient use of local representation and cannot aggregate local information from various domains.

In this work, we are motivated by the above observation. It is reasonable to assume that incorporating more local regions and information domains can improve detection performance. We expect to design a specialized model to implement this idea and verify its performance through extensive experiments. We aim to guide the model to capture subtle artifacts around some high-level facial local regions that are sensitive to forgeries due to complicated natural motions. These regions are referred to as the forgery-sensitive local regions (FSLR) in this paper, which are abundant in high-level semantics that can enhance the model's generalization capability. We also consider the feasibility of simultaneously exploiting information from space, frequency, and time domains based on a 3D CNN backbone.

To this end, we propose Cross-Domain Local Forensics (XDLF), a novel pipeline specially designed for feature extraction across multiple domains and local artifacts enhancement. Four forgery-sensitive local regions (i.e., left eye, right eye, nose, and mouth) are extracted to guide the model to capture subtle artifacts around these regions. To simultaneously leverage information from space, frequency, and time domains, we design a two-stream 3D CNN based framework to learn a cross-domain dense representation for forgery detection.

To demonstrate the effectiveness of our framework, extensive experiments were conducted on several benchmark datasets, including FaceForensics++~\cite{19ff}, Celeb-DF~\cite{20cdf}, and DFDC~\cite{20dfdc}. Our results show the superiority of the proposed method over many state-of-the-art approaches on cross-dataset generalization.

Our main contributions are as follows:

\begin{itemize}[topsep=0ex,parsep=.5ex]
    \item[$\bullet$] We leverage four forgery-sensitive local regions of a human face to guide the model to enhance subtle artifacts and localize potential anomalies around those regions. Using bounding boxes of those regions, we extract regional features as an attention to help the model focus more on those regions. We validated our design through ablations.
    
    \item[$\bullet$] We present a novel deepfake video detection pipeline that simultaneously exploits information from space, frequency, and time domains, thus learning a cross-domain dense representation for better generalization.
    
    \item[$\bullet$] We achieve impressive performance on extensive experiments, and our method outperforms several state-of-the-art methods on cross-dataset generalization.
\end{itemize}

%% file: contents/related-work.tex
\section{Related Work}

\subsection{Deepfake Detection}

Existing deepfake detection algorithms can fall into two categories, namely image-based methods and video-based methods, depending on whether temporal information is explicitly exploited across frames.

\noindent\textbf{Image-based Methods.}
Earlier image-based methods employed hand-crafted facial features to detect forgeries, e.g., steganalysis features~\cite{17twostream}, inconsistent head poses~\cite{19headpose}, and anomalous visual artifacts~\cite{19visualartifacts}. However, these methods underperformed on more realistic forgeries synthesized with more advanced face manipulation technologies recently. With the tremendous progress of deep learning, many works~\cite{18mesonet,19ff} utilized state-of-the-art convolutional neural networks (CNN), e.g., Xception~\cite{17xception}, to extract features from facial images and perform binary classification. More recently, an increasing number of CNN-based methods have been proposed from various perspectives. They aimed at exploring the crucial discrepancies between real and fake images, continuously improving the detection performance. These methods included leveraging frequency spectrum~\cite{20frank,20f3net,20twobranch,21hff,21spsl}, attention mechanism~\cite{20attention,21multiattention}, extra identity information~\cite{19obama,21idreveal}, self-supervised learning~\cite{19fwa,20facexray,21pcl}, etc.

\noindent\textbf{Video-based Methods.}
Unlike image-based methods, video-based methods distinguish real and fake videos based on a sequence of aligned frames. Most works managed to model the temporal consistency across frames, since current face manipulation techniques struggled to generate temporally coherent fake videos. These methods~\cite{18rnn,19cnngru,20twobranch,20cnn3d} utilized recurrent neural networks (RNN) or 3D CNN to extract spatio-temporal features of facial movements. They can focus on unnatural eye blinking~\cite{18eyeblink}, irregular mouth motion~\cite{21lipforensics}, inconsistent  visual-auditory modalities~\cite{20pvmismatch,20notmade,20emotion,21jointav}. In contrast, our method designs a two-stream 3D CNN based framework to mine forgery patterns from space, frequency, and time domains. We also leverage four facial forgery-sensitive local regions to enhance imperceptible artifacts for forgery defect localization.

\subsection{Generalization to Unseen Forgeries}

While current methods achieved excellent accuracy in the scenario where the training and test sets have similar distributions, they cannot generalize very well to unseen forgeries and tend to overfit to manipulation-specific artifacts. It is of paramount importance for deployed detectors to learn generalized representation regardless of forgery types. To this end, many works focused on improving generalization to unseen forgeries with diverse approaches. Several works~\cite{20twobranch,21hff} used a two-branch architecture to exploit information from the RGB domain and the frequency domain, exploiting generalized frequency patterns to expose the discrepancies. Our method has a similar idea but far different designs. Moreover, a series of self-supervised methods~\cite{19fwa,20facexray,21pcl} demonstrated superior generalization. These methods relied on self-generated fake data targeted at specific patterns without the need for conventional forgery training data. The patterns can be face warping artifacts~\cite{19fwa}, blending boundary~\cite{20facexray}, source feature inconsistency~\cite{21pcl}. LipForensics~\cite{21lipforensics} exhibited remarkable performance on cross-dataset generalization by pre-training a spatio-temporal network to perform lipreading and fine-tuning on a deepfake dataset. We followed its experimental settings due to similar goals.

%% file: contents/method.tex
\section{Proposed Method}

\subsection{Overview}
In this section, we first explain the motivation of our work, and then briefly introduce the pipeline of our proposed method.

\noindent\textbf{Motivation.}
Recent studies~\cite{19visualartifacts,20f3net,21lipforensics,21multiattention} have shown that the discrepancies between real and fake videos contain implicitly in local subtle regions, where manipulation artifacts may exist across various information domains. Unfortunately, most deepfake datasets have no manipulation masks as local supervision. Without external location guidance of facial semantic regions that are sensitive to forgeries, it is often difficult for detectors to localize those subtle artifacts. We observe that current detection algorithms had two limitations in leveraging local information:

\begin{itemize}[topsep=1ex,parsep=1ex]
    \item[$\bullet$] Some algorithms~\cite{18eyeblink,21lipforensics} relied on a single facial region as the criterion to detect forgeries, while ignoring the effect of other critical local regions, which may restrict the performance. Our framework exploits four forgery-sensitive local regions (FSLR) of a human face, which are used to guide the model to enhance subtle artifacts and localize more potential anomalies based on our newly proposed FSLR-Guided Feature Enhancement.
    
    \item[$\bullet$] Many algorithms made insufficient use of local regions to detect anomalies, which can be embodied in multiple information domains, e.g., space, frequency, and time domains. To the best of our knowledge, few studies have been done to simultaneously capture local features across these three domains. We note that the Two-branch~\cite{20twobranch} method extracted spatial/frequency and temporal features at two stages with CNN and RNN, respectively, without cross reference among these features. To this end, we propose a two-stream framework, Cross-Domain Local Forensics, to simultaneously exploit local information from those three domains.
\end{itemize}

\begin{figure}[t]
\centering
\includegraphics[width=\linewidth]{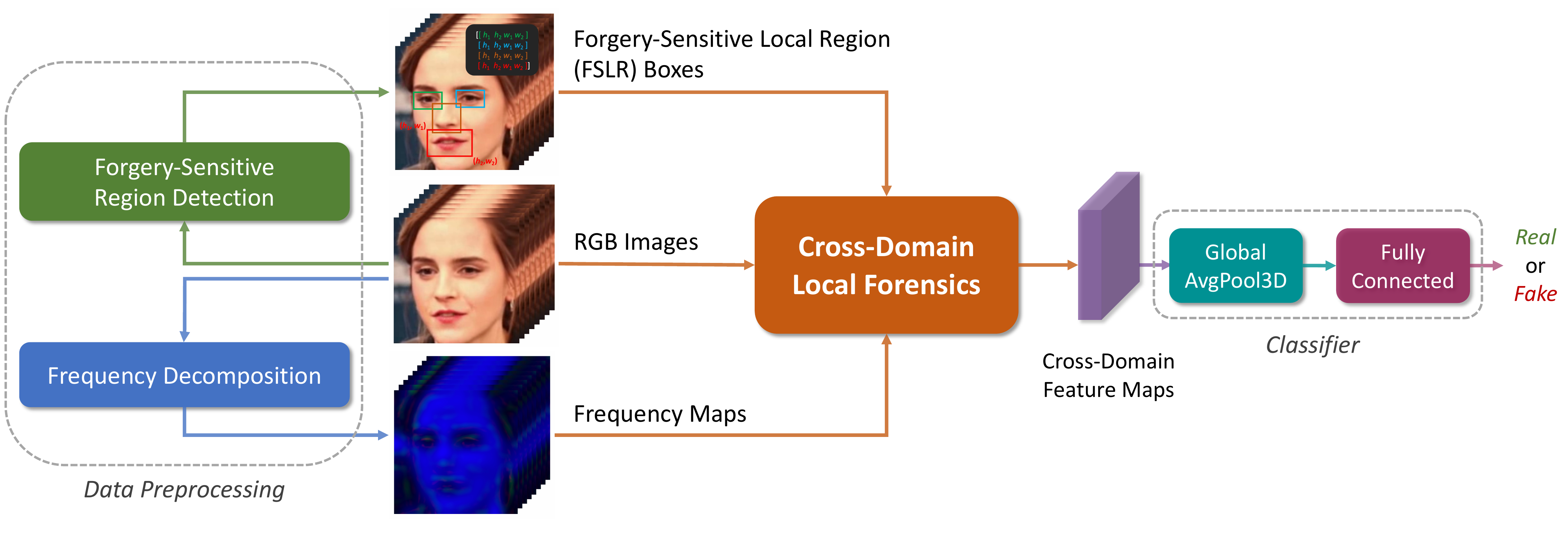}
\caption{Pipeline of our proposed framework XDLF. The end-to-end training consists of three stages: Data Preprocessing, Cross-Domain Local Forensics, and Classifier.}
\label{fig:xdlf_pipe}
\end{figure}

\noindent\textbf{Pipeline.}
Motivated by the above observations, we propose a novel feature extraction framework \textbf{Cross-Domain Local Forensics} (\textbf{XDLF}) for more general deepfake video detection. Fig.~\ref{fig:xdlf_pipe} illustrates the overall pipeline of XDLF. The pipeline takes as input a sequence of aligned RGB frames. First, the data preprocessing consists of two procedures. On the one hand, \textbf{Frequency Decomposition} takes as input RGB images to generate frequency maps where manipulation traces in the frequency domain are amplified, especially for those videos with high compression. On the other hand, \textbf{Forgery-Sensitive Region Detection} takes as input RGB images to extract bounding boxes of four \textbf{forgery-sensitive local regions} (\textbf{FSLR}) that are abundant in high-level defects. The four FSLRs are left eye, right eye, nose, and mouth. Then, sequences of RGB images, frequency maps, and FSLR boxes are input into \textbf{Cross-Domain Local Forensics} (\textbf{XDLF}) to learn a comprehensive and generalized cross-domain features. Finally, a classifier comprising a 3D global average pooling layer and a fully-connected layer is used to make predictions.

\subsection{Data Preprocessing}

\begin{figure}[b]
\centering
\includegraphics[width=.95\linewidth]{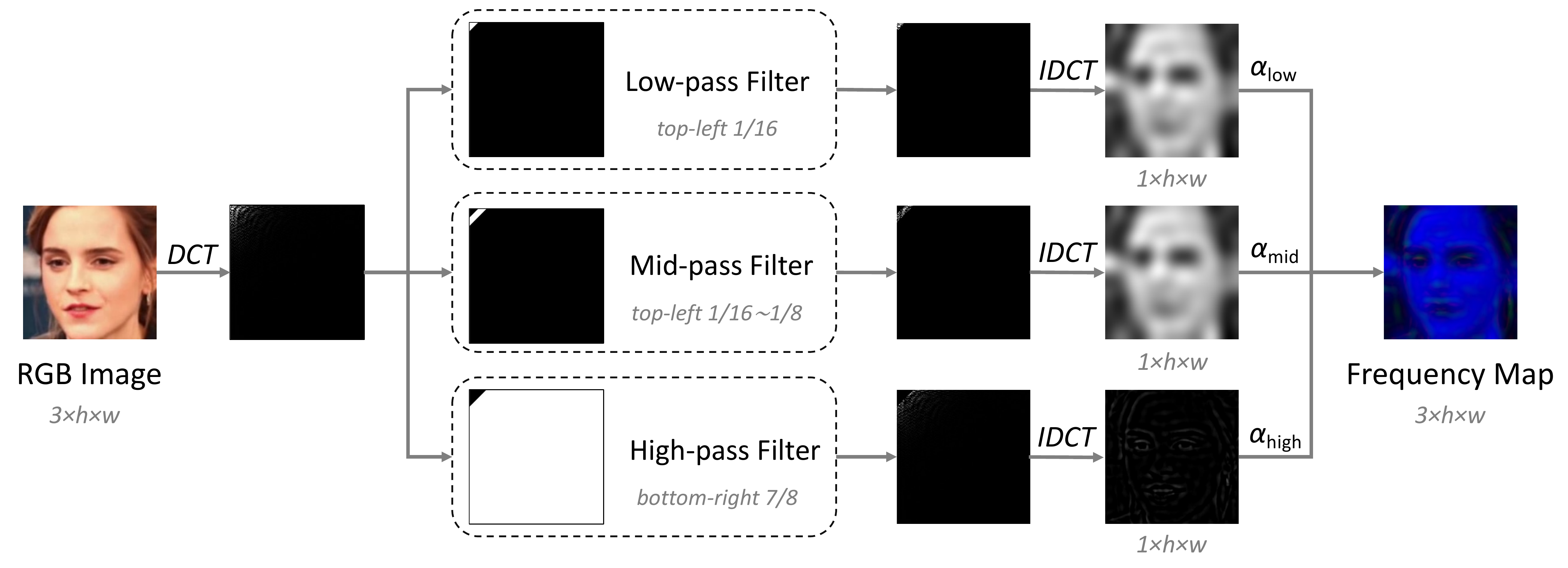} 
\caption{Pipeline of Frequency Decomposition. This module generates frequency maps where manipulation traces in the frequency domain are amplified adaptively.}
\label{fig:fd}
\end{figure}

\noindent\textbf{Frequency Decomposition.}
Recent studies~\cite{19zhang,21spsl} observed that up-sampling is a necessary step of most existing face manipulation methods, and cumulative up-sampling can leave apparent anomalies in the frequency domain, which provides clues for detecting manipulated videos. Inspired by F$^3$-Net~\cite{20f3net}, we design Frequency Decomposition to obtain multi-band frequency maps adaptively. Fig.~\ref{fig:fd} shows the pipeline of this module.

For each RGB image $\boldsymbol{X}$ in a frame sequence, we first calculate the frequency response with Discrete Cosine Transform (DCT) $\mathcal{D}$. Then, filters of low, middle, and high frequency bands $\boldsymbol{f_i},\;i\in\{\text{low},\text{mid},\text{high}\}$ are used to obtain three frequency components. We follow the settings in~\cite{20f3net} to construct filters. Next, Inversed Discrete Cosine Transform (IDCT) $\mathcal{D}^{-1}$ is applied to three frequency components to obtain the corresponding spatial components $\boldsymbol{Y_i},\;i\in\{\text{low},\text{mid},\text{high}\}$. Finally, the three spatial components are concatenated to attain the frequency map $\boldsymbol{Y}$. Before concatenation, each component is multiplied by a learnable weight $\alpha_i\in(0,1),\;i\in\{\text{low},\text{mid},\text{high}\}$ to enable the model to adaptively concentrate on the interested frequency band for a flexible representation of frequency features. The above can be summarized as Eq.~\ref{equ:fd},~\ref{equ:concat}, where $\odot$ is the element-wise product.
\begin{equation}
\label{equ:fd}
    \boldsymbol{Y_i} = \mathcal{D}^{-1}\{\mathcal{D}(\boldsymbol{X})\odot\boldsymbol{f_i}\},\;i\in\{\text{low},\text{mid},\text{high}\}
\end{equation}
\begin{equation}
\label{equ:concat}
    \boldsymbol{Y} = \text{Concat}(\alpha_\text{low}\boldsymbol{Y_\text{low}},\; \alpha_\text{mid}\boldsymbol{Y_\text{mid}},\; \alpha_\text{high}\boldsymbol{Y_\text{high}})
\end{equation}

\noindent\textbf{Forgery-Sensitive Region Detection.}
Current face manipulation techniques struggled to generate temporally coherent fake faces, especially in high-level semantic regions that have continual motions and thereby sensitive to forgeries. We hope to guide the model to pay more attention to these regions. Therefore, we extract bounding boxes of four forgery-sensitive local regions (FSLR): left eye, right eye, nose, and mouth. These four manually selected regions are further leveraged by \textbf{FSLR-Guided Feature Enhancement} (\textbf{FGFE}) as an external guidance. For each RGB image, we first compute 68 facial landmarks based on a face detector. Then, the landmarks are used to crop bounding boxes of those four regions based on preset box sizes. Each box can be expressed as a quadruple $(h_1,h_2,w_1,w_2)$ where $(h_1,w_1)$ is the top-left vertex and $(h_2,w_2)$ is the bottom-right vertex. The four boxes are stacked to generate the $4\times 4$ FSLR box matrix.

\subsection{Cross-Domain Local Forensics}

\begin{figure}[t]
\centering
\includegraphics[width=\linewidth]{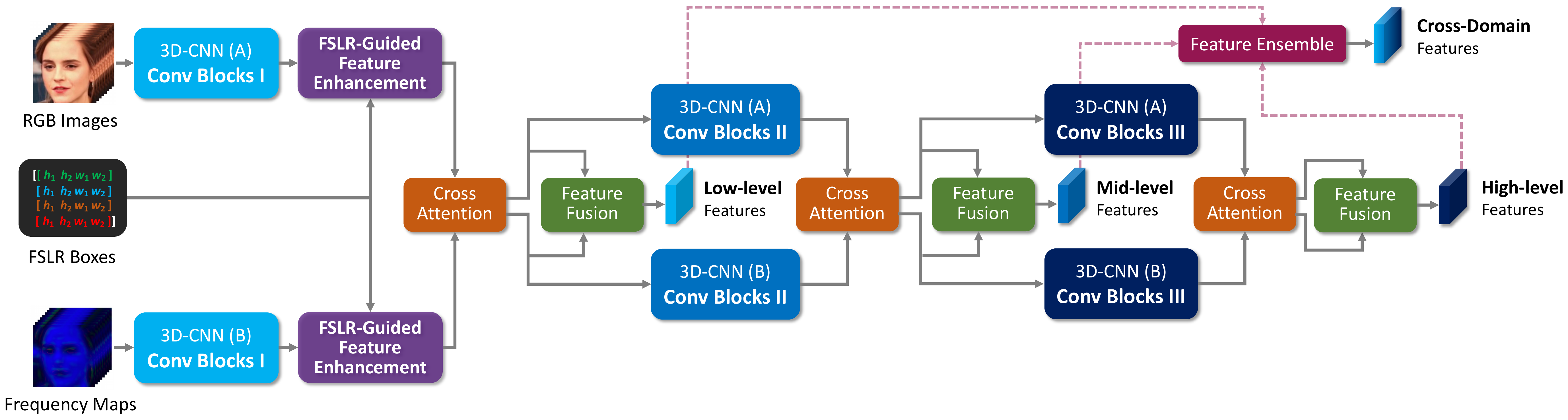} 
\caption{Framework of Cross-Domain Local Forensics. We adopt a two-stream architecture for cross-domain feature extraction based on two symmetric spatio-temporal convolutional backbones, e.g., 3D ResNet-50~\cite{16resnet,18resnet3d}.}
\label{fig:xdlf_fram}
\end{figure}

We propose a novel two-stream collaborative learning framework for cross-domain feature extraction, Cross-Domain Local Forensics (XDLF), which is based on a spatio-temporal convolutional backbone. As is illustrated in Fig.~\ref{fig:xdlf_fram}, the framework consists of two symmetric 3D CNN backbones: 3D-CNN(A) extracts spatio-temporal features of RGB images, and 3D-CNN(B) extracts frequency-temporal features of frequency maps. The features of the two modalities are cross-referenced and merged at low, middle, and high levels of the backbone, with \textbf{Cross Attention} and \textbf{Feature Fusion}, respectively. Moreover, we apply \textbf{FSLR-Guided Feature Enhancement} to the low-level features of both streams, thus enhancing the local subtle artifacts of shallow features under the guidance of forgery-sensitive regions. The ultimate cross-domain features are obtained with \textbf{Feature Ensemble} to integrate features of three different levels of abstraction.

\begin{figure}[b]
\centering
\includegraphics[width=.9\linewidth]{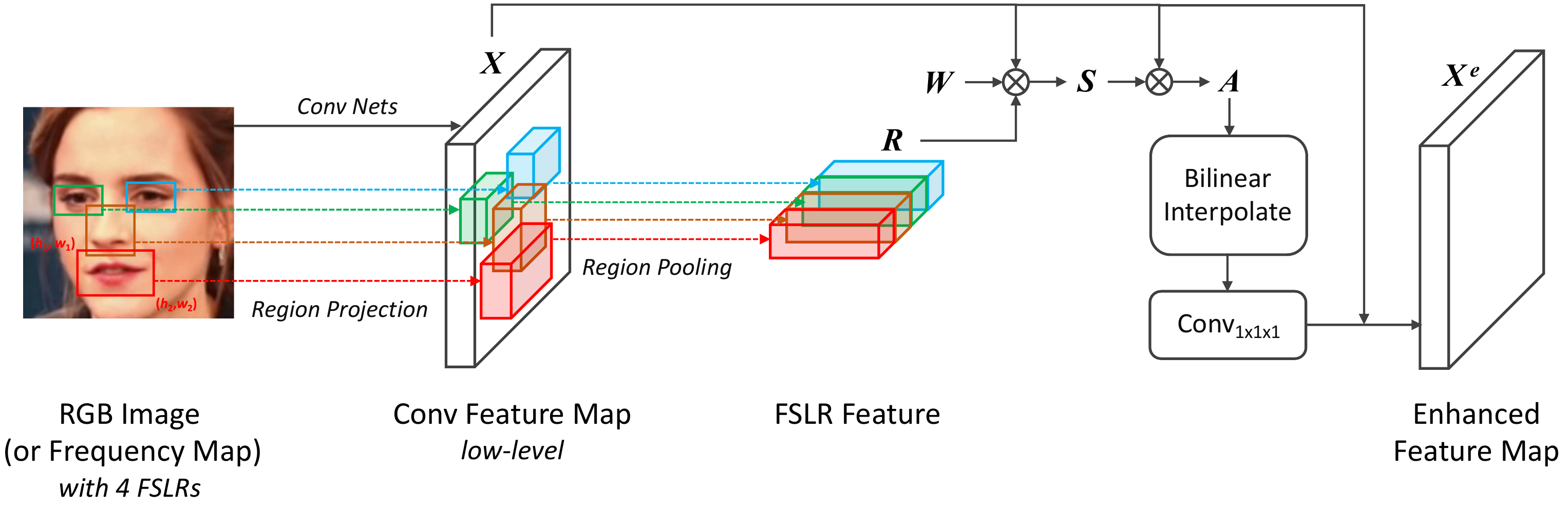} 
 \caption{Structure of FSLR-Guided Feature Enhancement. This module is designed to guide the model to enhance subtle artifacts of shallow features and localize more anomalous regions.}
\label{fig:fgfe}
\end{figure}

\noindent\textbf{FSLR-Guided Feature Enhancement.}
Many studies~\cite{21multiattention,21spsl} showed that local textural artifacts represent the high frequency component of shallow features, which is essential for the face forgery detection task. These artifacts are especially salient nearby critical facial regions that are sensitive to forgeries. As aforementioned, we exploit four forgery-sensitive local regions to enhance subtle artifacts and guide the model to localize more possible anomalies in these regions. The module structure is shown in Fig.~\ref{fig:fgfe}.

The module takes as input low-level RGB (or frequency) features $\boldsymbol{X}\in\mathbb{R}^{c\times d\times h\times w}$ (of $c$ channels, depth $d$, height $h$, width $w$) and FSLR boxes $\boldsymbol{r}\in\mathbb{Z}^{d\times 4\times 4}$ and outputs the enhanced features of the same shape. First, the region coordinates are scaled down (i.e., region projection) according to the size difference between the RGB image (or frequency map) and low-level features. Then, FSLR features $\boldsymbol{R}\in\mathbb{R}^{4\times c\times d\times H\times W}$ are obtained with region pooling, which refers to ROI pooling~\cite{15rcnn} in object detection. Specifically, we crop four sub-features with region coordinates and generate four FSLR features of fixed size ($H\times W$) using adaptive max-pooling (Eq.~\ref{equ:roi_pool}). FSLR features condense the irregular semantics of local textural patterns in these four regions, which serve as an attention for global features. Next, transformed features $\boldsymbol{X'}\in\mathbb{R}^{c\times hw}$ are calculated by temporally averaging the RGB (or frequency) features $\boldsymbol{X}$ and flattening spatial dimensions. And features $\boldsymbol{R'}\in\mathbb{R}^{4c\times dHW}$ are also obtained by flattening the FSLR features $\boldsymbol{R}$. Later, the similarity matrix $\boldsymbol{S}\in\mathbb{R}^{hw\times dHW}$ between $\boldsymbol{X'}$ and $\boldsymbol{R'}$ (Eq.~\ref{equ:simi}) is computed, where $\boldsymbol{W}\in\mathbb{R}^{c\times 4c}$ is a learnable weight matrix. Each value in $\boldsymbol{S}$ represents the similarity between each row in $\boldsymbol{X'}^T$ and each column in $\boldsymbol{R'}$. By the similarity matrix, we model the internal relevance between those local regions for cross-region forgery mining. And then the attention matrix $\boldsymbol{A}\in\mathbb{R}^{c\times dHW}$ is calculated to enhance the original features (Eq.~\ref{equ:att}). Moreover, the upsampled $\boldsymbol{A'}\in\mathbb{R}^{c\times d\times h\times w}$ is obtained by reshaping, bilinear interpolation, and $1\times1\times1$ convolution (Eq.~\ref{equ:bilinear},~\ref{equ:convbnrelu}). Finally, the enhanced features $\boldsymbol{X^e}\in\mathbb{R}^{c\times d\times h\times w}$ are attained by element-wise product and residual addition (Eq.~\ref{equ:residual}). We apply this module to the low-level features of both streams, which enables the model to pay more attention to the regularity and consistency of local semantic regions.

\begin{minipage}[b]{.47\linewidth}
\centering

\begin{equation}
\label{equ:roi_pool}
\boldsymbol{R}=\text{AdaMaxPool}(\boldsymbol{X},\text{Proj}(\boldsymbol{r}))
\end{equation}
\begin{equation}
\label{equ:simi}
\boldsymbol{S}=\boldsymbol{X'}^T\boldsymbol{W}\boldsymbol{R'} 
\end{equation}
\begin{equation}
\label{equ:att}
\boldsymbol{A}=\boldsymbol{X'}\boldsymbol{S}
\end{equation}

\end{minipage}
\medskip
\begin{minipage}[b]{.47\linewidth}
\centering  

\begin{equation}
\label{equ:bilinear}
\boldsymbol{A'}=\text{BilinearInterpolate}(\boldsymbol{A})   
\end{equation}
\begin{equation}
\label{equ:convbnrelu}
\boldsymbol{A'}=\text{ReLU}(\text{BN}(\text{Conv}_1(\boldsymbol{A'})))
\end{equation}
\begin{equation}
\label{equ:residual}
\boldsymbol{X^e}=\boldsymbol{X}+\boldsymbol{X}\odot\boldsymbol{A'}
\end{equation}

\end{minipage}

\noindent\textbf{Cross Attention.}
In this module, RGB and frequency features are cross-referenced at low, middle, and high levels of the backbone, which enables the model to learn a more comprehensive cross-domain representation. The module takes as input RGB features $\boldsymbol{X}$ and frequency features $\boldsymbol{X_f}$. First, the two features are concatenated on the channel axis and then applied $1\times1\times1$ convolution (Eq.~\ref{equ:ca_concat},~\ref{equ:ca_concat_2}). Next, $3\times3\times3$ convolution with output channel 2 and sigmoid activation is used to obtain two attention maps (Eq.~\ref{equ:conv3x3}). Finally, the original features are enhanced with attention maps by element-wise product (Eq.~\ref{equ:element_wise}).

\begin{minipage}[b]{.47\linewidth}
\centering

\begin{equation}
\label{equ:ca_concat}
\boldsymbol{U}=\text{Concat}(\boldsymbol{X},\boldsymbol{X_f})
\end{equation}
\begin{equation}
\label{equ:ca_concat_2}
\boldsymbol{U'}=\text{ReLU}(\text{BN}(\text{Conv}_1(\boldsymbol{U})))   
\end{equation}

\end{minipage}
\medskip
\begin{minipage}[b]{.47\linewidth}
\centering  

\begin{equation}
\label{equ:conv3x3}
\boldsymbol{A},\boldsymbol{A_f}=\text{Sigmoid}(\text{Conv}_3(\boldsymbol{U'}))
\end{equation}
\begin{equation}
\label{equ:element_wise}
\boldsymbol{X^c}=\boldsymbol{X}\odot\boldsymbol{A},\;\boldsymbol{X_f^c}=\boldsymbol{X_f}\odot\boldsymbol{A_f}
\end{equation}

\end{minipage}

\begin{figure}[t]
\centering
\includegraphics[width=.95\linewidth]{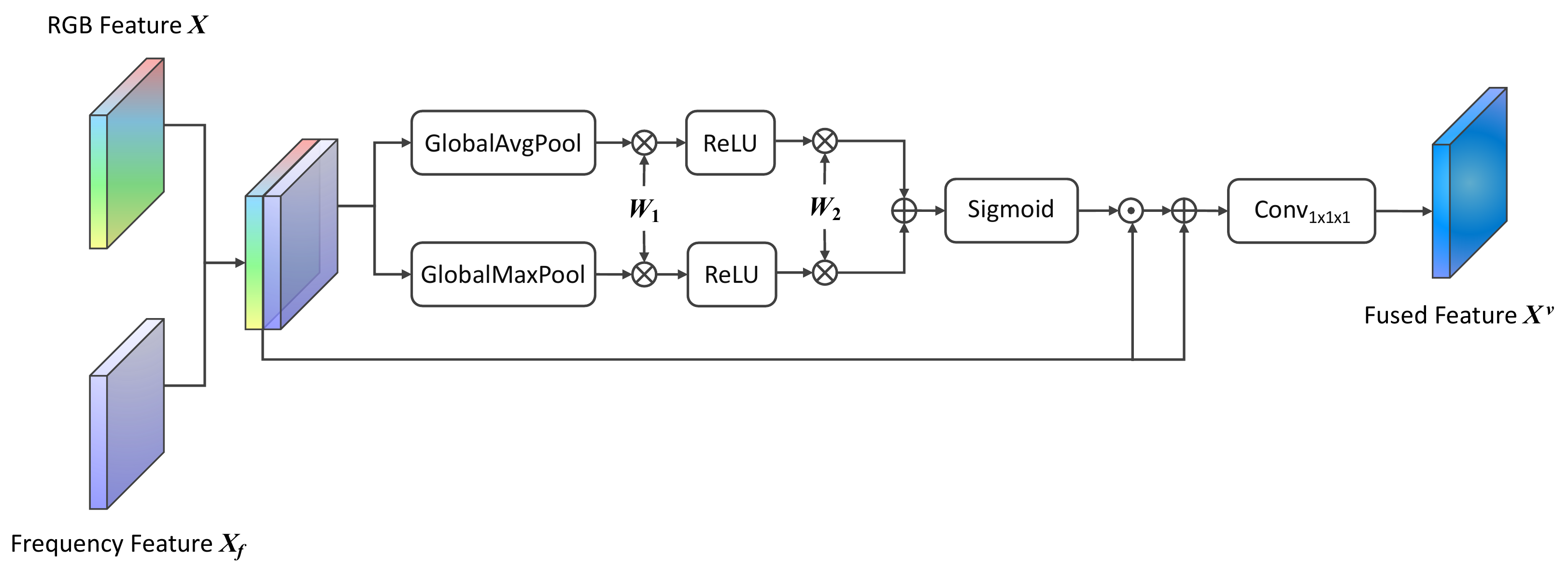} 
 \caption{Structure of Feature Fusion. This module is designed to model the interdependence between RGB and frequency features for improved cross-domain fusion.}
\label{fig:fusion}
\end{figure}

\noindent\textbf{Feature Fusion.}
In this module, RGB and frequency features are fused in a complementary way based on Squeeze-and-Excitation (SE)~\cite{18se}. SE block improves the quality of cross-domain features by explicitly modeling the interdependence between the channels of RGB and frequency features. The module structure is shown in Fig.~\ref{fig:fusion}.

This module also takes as input RGB features $\boldsymbol{X}\in\mathbb{R}^{C\times D\times H\times W}$ and frequency features $\boldsymbol{X_f}\in\mathbb{R}^{C\times D\times H\times W}$. The two features are first concatenated to obtain $\boldsymbol{U}\in\mathbb{R}^{2C\times D\times H\times W}$ (Eq.~\ref{equ:ff_concat}). Then, the spatial information is squeezed into a value by global pooling to get channel descriptor $\boldsymbol{V}\in\mathbb{R}^{2C}$ (Eq.~\ref{equ:globalpool},~\ref{equ:globalpool_2}). Next, we enable channel descriptor $\boldsymbol{V}$ to capture the interdependency between channels and obtain channel attention $\boldsymbol{A_c}\in\mathbb{R}^{2C}$ (Eq.~\ref{equ:channel_att},~\ref{equ:channel_att_2},~\ref{equ:channel_att_3}), where $\boldsymbol{W_1}\in\mathbb{R}^{2C\times\frac{2C}{r}}$ and $\boldsymbol{W_2}\in\mathbb{R}^{\frac{2C}{r}\times 2C}$ are learnable weight matrices, $r$ is the reduction ratio. Finally, the fused features $\boldsymbol{X^v}$ are computed as Eq.~\ref{equ:ff_residual}.

\begin{minipage}[b]{.47\linewidth}
\centering

\begin{equation}
\label{equ:ff_concat}
\boldsymbol{U}=\text{Concat}(\boldsymbol{X},\boldsymbol{X_f})
\end{equation}
\begin{equation}
\label{equ:globalpool}
\boldsymbol{V}_\text{avg}=\text{GlobalAvgPool}(\boldsymbol{U})
\end{equation}
\begin{equation}
\label{equ:globalpool_2}
\boldsymbol{V}_\text{max}=\text{GlobalMaxPool}(\boldsymbol{U})
\end{equation}

\end{minipage}
\medskip
\begin{minipage}[b]{.47\linewidth}
\centering

\begin{equation}
\label{equ:channel_att}
\boldsymbol{V'_\text{avg}}=\boldsymbol{W_2}\text{ReLU}(\boldsymbol{W_1}\boldsymbol{V_\text{avg}})    
\end{equation}
\begin{equation}
\label{equ:channel_att_2}
\boldsymbol{V'_\text{max}}=\boldsymbol{W_2}\text{ReLU}(\boldsymbol{W_1}\boldsymbol{V_\text{max}})     
\end{equation}
\begin{equation}
\label{equ:channel_att_3}
\boldsymbol{A_c}=\text{Sigmoid}(\boldsymbol{V'_\text{avg}}+\boldsymbol{V'_\text{max}})  
\end{equation}

\end{minipage}
\begin{equation}
\label{equ:ff_residual}
\boldsymbol{X^v}=\text{ReLU}(\text{BN}(\text{Conv}_{1\times1\times1}(\boldsymbol{U}+\boldsymbol{U}\odot\boldsymbol{A_c})))
\end{equation}

\noindent\textbf{Feature Ensemble.}
This module aggregates low, middle, and high-level features through adaptive average pooling and concatenation (Eq.~\ref{equ:ens_1},~\ref{equ:ens_2}, \ref{equ:ens_3},~\ref{equ:ens_4}).
\begin{align}
\label{equ:ens_1}
\widetilde{\boldsymbol{X}}^\text{low} &= \lambda_\text{low}\text{AdaAvgPool}(\boldsymbol{X}^\text{low}, (d_\text{high},h_\text{high},w_\text{high})) \\
\label{equ:ens_2}
\widetilde{\boldsymbol{X}}^\text{mid} &= \lambda_\text{mid}\text{AdaAvgPool}(\boldsymbol{X}^\text{mid}, (d_\text{high},h_\text{high},w_\text{high})) \\
\label{equ:ens_3}
\widetilde{\boldsymbol{X}}^\text{high} &= \lambda_\text{high}\boldsymbol{X}^\text{high} \\
\label{equ:ens_4}
\widetilde{\boldsymbol{X}} &= \text{Concat}(\widetilde{\boldsymbol{X}}^\text{low},\widetilde{\boldsymbol{X}}^\text{mid},\widetilde{\boldsymbol{X}}^\text{high})
\end{align}
where $\boldsymbol{X}^i\in\mathbb{R}^{c_i\times d_i\times h_i\times w_i},\;i\in\{\text{low},\text{mid},\text{high}\}$ are fused features of three abstraction levels, and $\lambda_\text{i}\in(0,1),\;i\in\{\text{low},\text{mid},\text{high}\}$ are three learnable parameters for adaptive feature combination.

%% file: contents/experiment.tex
\section{Experiment and Discussion}

\subsection{Experiment Setup}

\noindent\textbf{Datasets.}
We used \textbf{FaceForensics++} (FF++)~\cite{19ff} for training and validation, and evaluated the cross-dataset generalization on \textbf{Celeb-DF} (CDF)~\cite{20cdf} and \textbf{DeepFake Detection Challenge} (DFDC)~\cite{20dfdc}. (1) FF++ is the most commonly used benchmark dataset containing 1,000 real videos and 4,000 fake videos. Each real video is manipulated by four face forgery techniques, i.e., DeepFakes (DF)~\cite{deepfakes}, FaceSwap (FS)~\cite{faceswap}, Face2Face (F2F)~\cite{16face2face}, and NeuralTextures (NT)~\cite{19neuraltextures}. We adopted the slightly-compressed (HQ/c23) and heavily-compressed (LQ/c40) versions for our experiments. (2) CDF is a challenging dataset that includes 590 real videos and 5,639 fake videos synthesized by an improved algorithm. (3) DFDC is a large-scale dataset with extreme filming conditions and various perturbations, which is also very challenging for current deepfake detectors. We used the preview version~\cite{19dfdcp} that includes 1,131 real videos and 4,113 fake videos for our evaluation.

\noindent\textbf{Evaluation Metrics.}
Following most existing works~\cite{20twobranch,20facexray,21lipforensics}, Accuracy (ACC) and Area Under the Receiver Operating Characteristic Curve (AUC) were used as the metrics to evaluate our method. As in~\cite{21lipforensics}, we reported video-level metrics for fair comparison with image-based methods. Specifically, all frame/clip predictions were averaged across the video and hence all models predicted based on an equal number of frames.

\noindent\textbf{Implementation Details.}
For each video, we sampled non-overlapping frame clips with a length of 16, and oversampled the minority class (e.g., real in FF++) to tackle label imbalance. We used the state-of-the-art face detector RetinaFace~\cite{20retina} to crop facial images with a size of $224\times224$ and FSLR box matrices with a size of $4\times4$. The preset FSLR size is $40\times40$ for the mouth and $30\times30$ for the other three. For data augmentations, we applied several traditional image augmentations such as random horizontal flipping. Moreover, as in~\cite{ntechlab}, we conducted Mixup~\cite{18mixup} augmentation on aligned real-fake pairs to reduce overfitting. For XDLF, we adopted 3D ResNet-50~\cite{16resnet,18resnet3d} as the backbone which is pre-trained on large-scale action recognition datasets to accelerate the model convergence. For FSLR-Guided feature enhancement, we set FSLR feature size $H=W=7$. For feature fusion, we set reduction ratio $r=16$. For training, we used a batch size of 4 and AdamW~\cite{19adamw} optimizer with initial learning rate $1\times10^{-4}$ and weight decay $1\times10^{-4}$. The learning rate decayed with a cosine annealing~\cite{17cosine} strategy with $T_\text{max}=32$. 

\subsection{In-dataset Evaluation}

We evaluated our method in the in-dataset scenario where the training and test sets have identical distributions. Following~\cite{21lipforensics}, we compared our method with current state-of-the-art approaches in FF++ under different quality settings (HQ/LQ). As shown in Table~\ref{tab:in_dataset}, we achieve great improvements over most current methods, especially under the challenging low-quality (LQ) setting where frequency statistics are partly destroyed. However, our method still maintains good performance when exploiting frequency spectrum, which we attribute to our two-stream architecture that learns to be biased towards RGB features. Note that we gain comparable results with LipForensics~\cite{21lipforensics}, which leverages dynamic lip features from pre-trained lipreading models.  Unlike LipForensics, our method does not need any external pre-training data and can be more efficiently trained.

\input{tables/in-dataset}

\begin{figure}[H]
\centering
\subfigure[LipForensics]{\includegraphics[width=.4\linewidth]{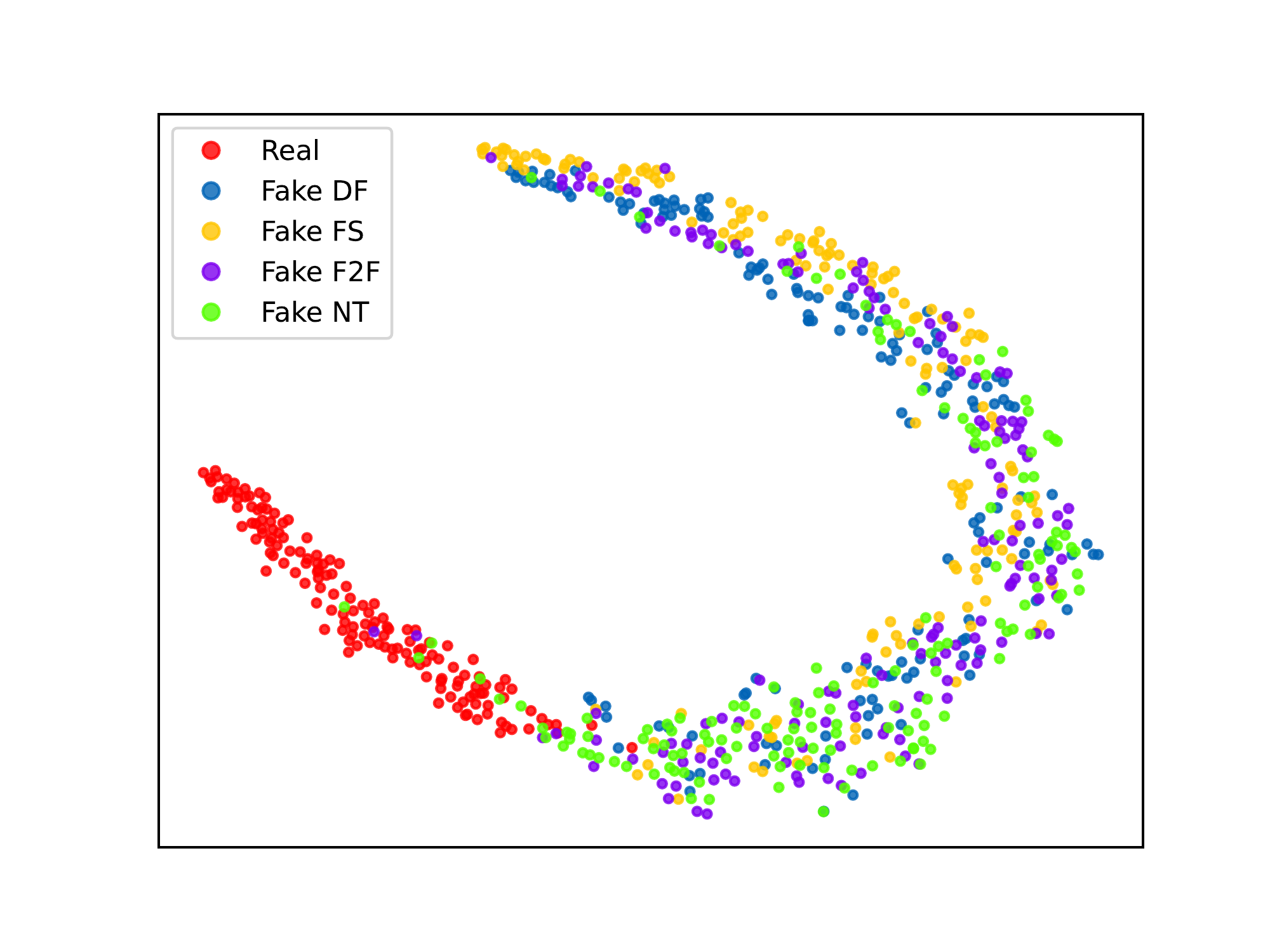}}
\qquad
\subfigure[XDLF]{\includegraphics[width=.4\linewidth]{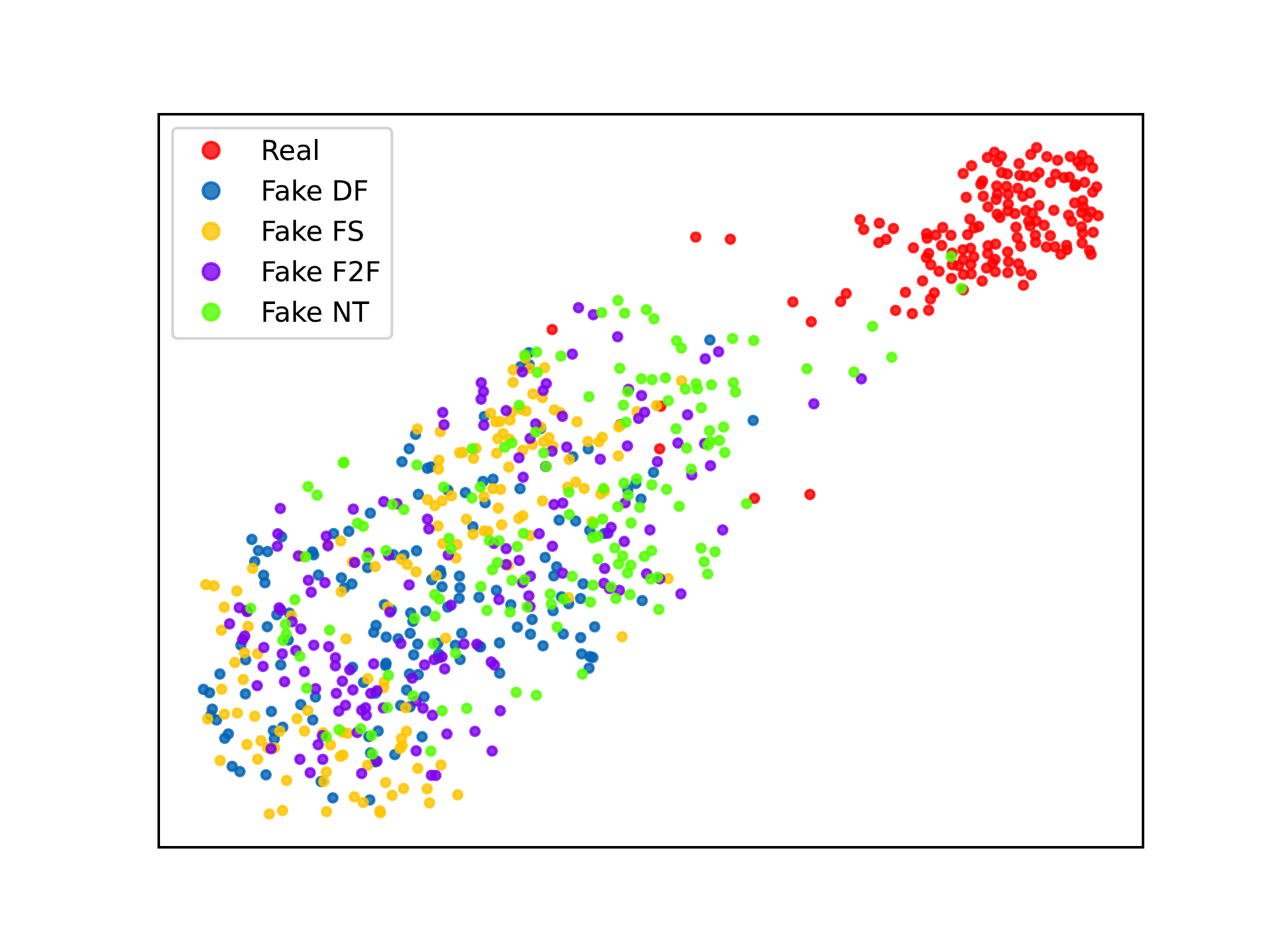}}
\caption{The t-SNE feature visualization of the baseline LipForensics~\cite{21lipforensics} (a) and our proposed XDLF (b) on FF++(HQ) test set. Each dot represents the feature of a video clip. Red dots are real clips, while the rest are fake ones with different forgery types.}
\label{fig:tsne}
\end{figure}

Moreover, we show the t-SNE~\cite{tsne} visualization of features extracted from classifiers of LipForensics and our method on FF++ high-quality (HQ) test set in Fig.~\ref{fig:tsne}. We observe that although both methods can well distinguish real and fake data, they learn different feature distributions. For LipForensics, the separation distances between real and fake data are smaller than our method, which can easily lead to classification ambiguity in those in-between videos, especially for some real and NeuralTextures-based fake samples. On the other hand, our method learns a more mixed and gathered feature representation of FF++ fake data without obviously separating different forgery types. It proves that our method can learn a generalized feature to detect novel forgeries.

\subsection{Cross-dataset Evaluation}
\label{sec:cross_dataset}

In real-world scenarios, a deployed detector is expected to identify videos crafted by unseen manipulations with unknown source videos. Therefore, we conducted cross-dataset evaluation as in~\cite{21lipforensics} to verify the generalization capability of our method. Specifically, we trained the models on FF++(HQ) and tested them on CDF and DFDC. As shown in Table~\ref{tab:cross_dataset}, our method outperforms all listed methods on both unseen datasets, surpassing the recent state-of-the-art LipForensics~\cite{21lipforensics} by 0.2\% and 0.3\% in terms of AUC on CDF and DFDC, respectively.

\input{tables/cross-dataset}

\subsection{Ablation Study}

\input{tables/module-ablations}

\noindent\textbf{Evaluations on Core Modules in XDLF.}
To understand the components responsible for our method's performance, we ablated three core modules in XDLF and examined its in-dataset and cross-dataset generalization performance. The modules are FSLR-Guided Feature Enhancement (\textbf{FGFE}), Cross Attention, and Feature Fusion. For the first two, we removed them directly as their inputs and outputs have the same shapes. For Feature Fusion, we replaced it with a simple channel-axis concatenation of RGB and frequency features. We trained all the models on FF++(HQ) and tested them on FF++(HQ), CDF, and DFDC. 

The results are shown in Table~\ref{tab:module-ablations}. We have the following observations: (1) Training our model without FGFE leads to a performance drop on all datasets. In cross-dataset evaluation, the model decreases by 3.4\% and 4.0\% in terms of AUC on CDF and DFDC, respectively. This suggests that the model learns more generalized features by enhancing subtle artifacts in those forgery-sensitive local regions. (2) Both Cross Attention and Feature Fusion play an essential role in performance improvements. Although they have the same goal to complementarily exploit forgery patterns from different domains, they work differently and enhance the model's performance mutually.

\begin{figure}[H]
\centering
\includegraphics[width=.7\linewidth]{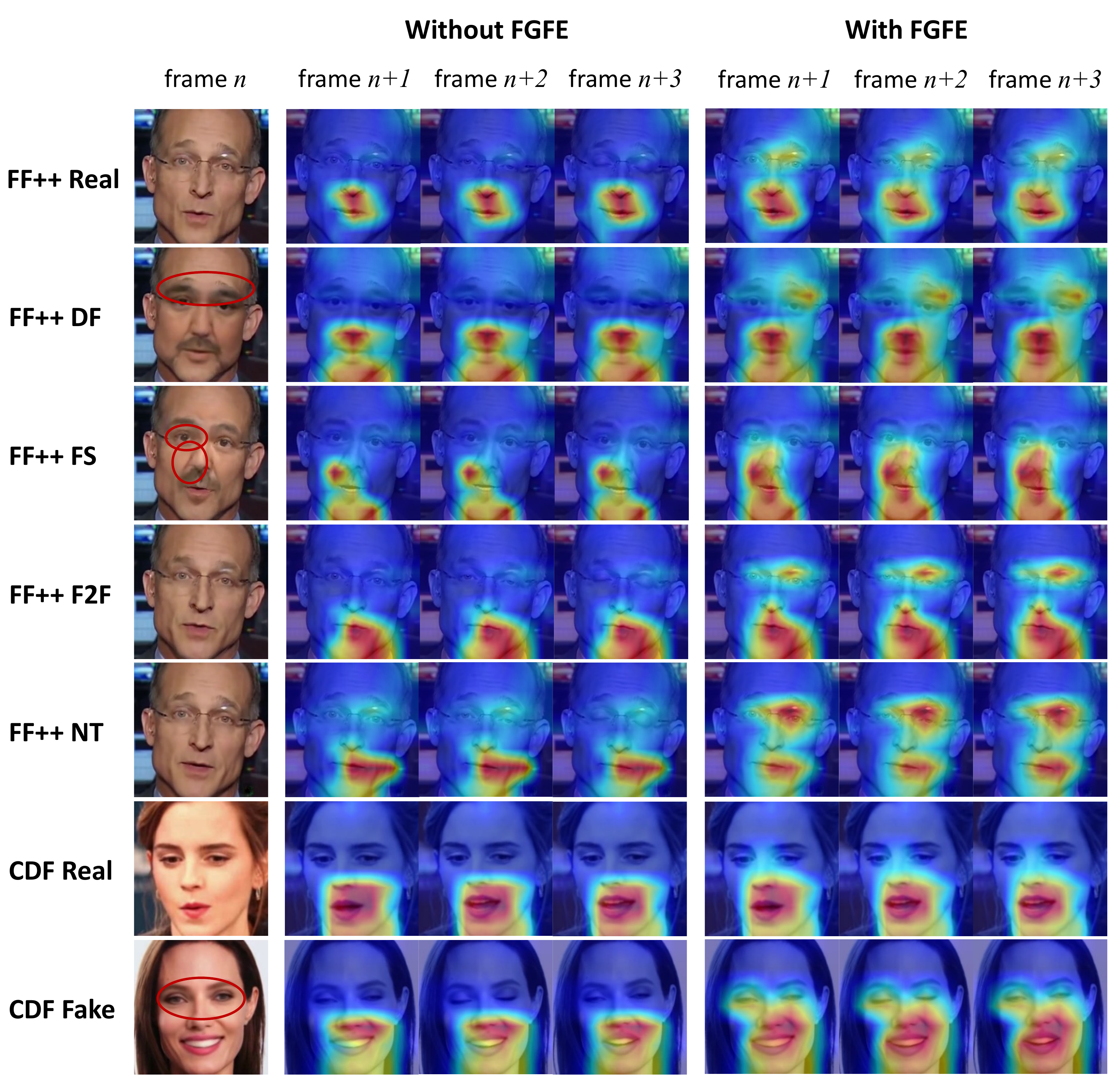}
\caption{The Grad-CAM visualization of localized defect regions of the model without/with FSLR-Guided Feature Enhancement (FGFE). We show several examples including four forgery types in FF++ and another dataset CDF. For each example, red circles mark visually noticeable artifacts, and consecutive frames in a video clip are provided to understand temporal defects. The warmer region suggests a higher probability of cross-domain defects the model believes.}
\label{fig:gradcam}
\end{figure}

To further understand the effect of FGFE, we show the Grad-CAM~\cite{17gradcam} visualization of the model without/with FGFE in Fig.~\ref{fig:gradcam}. It visually explains that FGFE serves as external guidance to help the model focus on four forgery-sensitive local regions. As can be seen, these regions are abundant in motions that contain more subtle artifacts. The model can localize more potential anomalies to detect forgeries with the help of FGFE, which is consistent with our motivation.

\input{tables/domain-ablations}

\noindent\textbf{Evaluations on Different Information Domains.}
We altered our feature extraction framework XDLF to prove the necessity to mine forgery clues from three different information domains, i.e., space, frequency, and time domains. Specifically, we trained three variants with each dropping one of the three domains: (1) \textbf{Freq-Freq-3D}: The inputs of both streams are the same frequency maps, and the network structure is unchanged. (2) \textbf{RGB-RGB-3D}: The inputs of both streams are RGB images, and the network structure is unchanged. (3) \textbf{RGB-Freq-2D}: The inputs are still RGB images and frequency maps, but temporal dimension is merged into batch dimension. We replaced the 3D ResNet-50 backbone with 2D ResNet-50 backbone, and replaced all 3D convolutional layers and 3D batch normalization layers with 2D counterparts.

The results are shown in Table~\ref{tab:domain-ablations}. We have the following observations: (1) By using 3D spatio-temporal CNN instead of 2D CNN, the in-dataset and cross-dataset generalization performance are all considerably improved. It indicates that our method can leverage 3D CNN to effectively capture temporal defects for forgery detection. (2) Compared to RGB-RGB-3D, Freq-Freq-3D achieves better cross-dataset generalization. It suggests that frequency statistics are more generalizable features than color textures. However, RGB-RGB-3D gains better in-dataset results which may benefit from manipulation-specific artifacts. (3) We note that forgery clues from these three domains work in a complementary way and contribute to the overall performance.

%% file: tables/in-dataset.tex
\setlength{\tabcolsep}{4pt}
\begin{table}[h]
    \caption{\textbf{In-dataset performance comparisons.} We report video-level ACC/AUC (\%) when trained and tested on FF++ slightly-compressed (HQ) and heavily-compressed (LQ) videos. The results of other methods are quoted from~\cite{21lipforensics}. The best results are in \textbf{bold}, and the second-best results are \underline{underlined}.}
    \centering
    \resizebox{.85\linewidth}{!}{
    \begin{tabular}{lccccc}
        \toprule
        \multirow{2}{*}{Method} & 
        \multicolumn{2}{c}{FF++(HQ)} & &
        \multicolumn{2}{c}{FF++(LQ)} \\
        \noalign{\smallskip}\cline{2-3}\cline{5-6}\noalign{\smallskip}
        & \quad ACC (\%) \quad & \quad AUC (\%) \quad &
        & \quad ACC (\%) \quad & \quad AUC (\%) \quad \\
        \noalign{\smallskip}\hline\noalign{\smallskip}
        Xception~\cite{19ff}
        & 97.0 & 99.3 & & 89.0 & 92.0 \\
        CNN-aug~\cite{20cnnaug}
        & 96.9 & 99.1 & & 81.9 & 86.9 \\
        Patch-based~\cite{20patchbased}
        & 92.6 & 97.2 & & 79.1 & 78.3 \\
        Two-branch~\cite{20twobranch}
        & -- & 99.1 & & -- & 91.1 \\
        Face X-ray~\cite{20facexray}
        & 78.4 & 97.8 & & 34.2 & 77.3 \\
        CNN-GRU~\cite{19cnngru}
        & 97.0 & 99.3 & & 90.1 & 92.2 \\
        LipForensics~\cite{21lipforensics}
        & \textbf{98.8} & \textbf{99.7} & & \underline{94.2} & \textbf{98.1} \\
        \noalign{\smallskip}\hline\noalign{\smallskip}
        XDLF (ours)
        & \underline{98.1} & \textbf{99.7} & & \textbf{94.5} & \underline{96.7} \\
        \bottomrule
    \end{tabular}
    }
    \label{tab:in_dataset}
\end{table}
\setlength{\tabcolsep}{1.4pt}

%% file: tables/cross-dataset.tex
\setlength{\tabcolsep}{4pt}
\begin{table}[t]
    \caption{\textbf{Cross-dataset performance comparisons.} We report video-level AUC (\%) on CDF and DFDC when trained on FF++(HQ). The results of other methods are quoted from~\cite{21lipforensics}. The best results are in \textbf{bold}.}
    \centering
    \resizebox{.64\linewidth}{!}{
    \begin{tabular}{lcc}
        \toprule
        Method &
        \quad CDF AUC (\%) \quad &
        \quad DFDC AUC (\%) \quad \\
        \noalign{\smallskip}\hline\noalign{\smallskip}
        Xception~\cite{19ff} & 73.7 & 70.9 \\
        CNN-aug~\cite{20cnnaug} & 75.6 & 72.1 \\
        Patch-based~\cite{20patchbased} & 69.6 & 65.6 \\
        Face X-ray~\cite{20facexray} & 79.5 & 65.5 \\
        CNN-GRU~\cite{19cnngru} & 69.8 & 68.9 \\
        Multi-task~\cite{19multitask} & 75.7 & 68.1 \\
        DSP-FWA~\cite{19fwa} & 69.5 & 67.3 \\
        Two-branch~\cite{20twobranch} & 76.7 & -- \\ 
        LipForensics~\cite{21lipforensics} & 82.4 & 73.5 \\
        \noalign{\smallskip}\hline\noalign{\smallskip}
        XDLF (ours) & \textbf{82.6} & \textbf{73.8} \\
        \bottomrule
    \end{tabular}
    }
    \label{tab:cross_dataset}
\end{table}
\setlength{\tabcolsep}{1.4pt}

%% file: tables/module-ablations.tex
\definecolor{mygray}{gray}{.9}

\setlength{\tabcolsep}{4pt}
\begin{table}[b]
    \caption{\textbf{Evaluations on core modules in XDLF.} We report video-level ACC/AUC (\%) on FF++(HQ), CDF, and DFDC when trained on FF++(HQ). The \colorbox{mygray}{highlighted} row is our original setting. We ablated core modules in our feature extraction framework to verify their effects. The best results are in \textbf{bold}.}
    \centering
    \resizebox{.73\linewidth}{!}{
    \begin{tabular}{lcccccccc}
    \toprule
    \multirow{2}{*}{Method} &
    \multicolumn{2}{c}{FF++(HQ)} & &
    \multicolumn{2}{c}{CDF} & &
    \multicolumn{2}{c}{DFDC} \\
    \noalign{\smallskip}\cline{2-3}\cline{5-6}\cline{8-9}\noalign{\smallskip}
    & ACC & AUC & & ACC & AUC & & ACC & AUC \\
    \noalign{\smallskip}\hline\noalign{\smallskip} \rowcolor{mygray}
    XDLF (ours)
    & \textbf{98.1} & \textbf{99.7} &
    & \textbf{74.2} & \textbf{82.6} &
    & \textbf{66.2} & \textbf{73.8} \\ 
    \quad w/o FGFE
    & 97.9 & 99.3 &
    & 71.7 & 79.2 &
    & 65.3 & 69.8 \\ 
    \quad w/o Cross Attention
    & 97.8 & 99.4 &
    & 72.2 & 79.9 &
    & 65.9 & 71.0 \\
    \quad w/o Feature Fusion
    & 98.0 & 99.4 &
    & 73.8 & 81.5 &
    & 66.1 & 72.3 \\ 
    \bottomrule
    \end{tabular}
    }
    \label{tab:module-ablations}
\end{table}
\setlength{\tabcolsep}{1.4pt}

%% file: tables/domain-ablations.tex
\definecolor{mygray}{gray}{.9}

\setlength{\tabcolsep}{4pt}
\begin{table}[b]
    \caption{\textbf{Evaluations on different information domains.} We report video-level ACC/AUC (\%) on FF++(HQ), CDF, and DFDC when trained on FF++(HQ). The \colorbox{mygray}{highlighted} row is our original setting. We developed three variants of feature extraction framework with each dropping one of the three domains, i.e., space, frequency, and time domains. The best results are in \textbf{bold}.}
    \centering
    \resizebox{\linewidth}{!}{
    \begin{tabular}{lcccccccccccc}
    \toprule
    \multirow{2}{*}{Method} &
    \multicolumn{3}{c}{Information Domains} & &
    \multicolumn{2}{c}{FF++(HQ)} & &
    \multicolumn{2}{c}{CDF} & &
    \multicolumn{2}{c}{DFDC} \\
    \noalign{\smallskip}\cline{2-4}\cline{6-7}\cline{9-10}\cline{12-13}\noalign{\smallskip}
    & Space & Frequency & Time &
    & ACC & AUC & & ACC & AUC & & ACC & AUC \\
    \noalign{\smallskip}\hline\noalign{\smallskip} \rowcolor{mygray}
    RGB-Freq-3D (ours) & $\checkmark$ & $\checkmark$ & $\checkmark$ &
    & \textbf{98.1} & \textbf{99.7} &
    & \textbf{74.2} & \textbf{82.6} &
    & \textbf{66.2} & \textbf{73.8} \\ 
    Freq-Freq-3D & $\times$ & $\checkmark$ & $\checkmark$ &
    & 97.6 & 99.0 &
    & 73.9 & \textbf{82.6} &
    & 64.5 & 72.5 \\
    RGB-RGB-3D & $\checkmark$ & $\times$ & $\checkmark$ &
    & \textbf{98.1} & 99.5 &
    & 72.7 & 81.2 &
    & 63.6 & 71.9 \\ 
    RGB-Freq-2D & $\checkmark$ & $\checkmark$ & $\times$ &
    & 96.4 & 98.5 &
    & 68.4 & 76.1 &
    & 61.3 & 69.1 \\ 
    \bottomrule
    \end{tabular}
    }
    \label{tab:domain-ablations}
\end{table}
\setlength{\tabcolsep}{1.4pt}

%% file: contents/conclusion.tex
\section{Conclusion}

In this paper, we propose Cross-Domain Local Forensics (XDLF), a specially designed pipeline for general deepfake video detection. Our approach aims at exploiting forgery patterns from space, frequency, and time domains simultaneously to learn a generalized cross-domain features. We also leverage four forgery-sensitive local regions to guide the model to capture subtle forgery defects. Experiments show that our method achieves impressive performance, especially strong cross-dataset generalization. We hope our work encourages future research on cross-domain forensics for more general deepfake detection.

\noindent\textbf{Acknowledgements.}
This work was supported by the National Natural Science Foundation of China (62271307, 61771310) and Key Lab of Information Network Security of Ministry of Public Security (The Third Research Institute of Ministry of Public Security). Shilin Wang is the corresponding author.

%% file: paper.bbl
\begin{thebibliography}{10}
\providecommand{\url}[1]{\texttt{#1}}
\providecommand{\urlprefix}{URL }
\providecommand{\doi}[1]{https://doi.org/#1}

\bibitem{18mesonet}
Afchar, D., Nozick, V., Yamagishi, J., Echizen, I.: Mesonet: a compact facial
  video forgery detection network. In: International Workshop on Information
  Forensics and Security (WIFS). pp.~1--7. IEEE (2018)

\bibitem{20pvmismatch}
Agarwal, S., Farid, H., Fried, O., Agrawala, M.: Detecting deep-fake videos
  from phoneme-viseme mismatches. In: Conference on Computer Vision and Pattern
  Recognition Workshops (CVPRW). pp. 2814--2822. IEEE/CVF (2020)

\bibitem{19obama}
Agarwal, S., Farid, H., Gu, Y., He, M., Nagano, K., Li, H.: Protecting world
  leaders against deep fakes. In: Conference on Computer Vision and Pattern
  Recognition Workshops (CVPRW). pp. 38--45. IEEE/CVF (2019)

\bibitem{19opticalflow}
Amerini, I., Galteri, L., Caldelli, R., Bimbo, A.D.: Deepfake video detection
  through optical flow based cnn. In: International Conference on Computer
  Vision Workshops (ICCVW). pp. 1205--1207. IEEE (2019)

\bibitem{20patchbased}
Chai, L., Bau, D., Lim, S.N., Isola, P.: What makes fake images detectable?
  understanding properties that generalize. In: European Conference on Computer
  Vision (ECCV). Lecture Notes in Computer Science, vol. 12371, pp. 103--120.
  Springer (2020)

\bibitem{21localrelation}
Chen, S., Yao, T., Chen, Y., Ding, S., Li, J., Ji, R.: Local relation learning
  for face forgery detection. In: AAAI Conference on Artificial Intelligence
  (AAAI). pp. 1081--1088. AAAI Press (2021)

\bibitem{17xception}
Chollet, F.: Xception: Deep learning with depthwise separable convolutions. In:
  Conference on Computer Vision and Pattern Recognition (CVPR). pp. 1800--1807.
  IEEE/CVF (2017)

\bibitem{20notmade}
Chugh, K., Gupta, P., Dhall, A., Subramanian, R.: Not made for each
  other-audio-visual dissonance-based deepfake detection and localization. In:
  ACM International Conference on Multimedia (ACM MM). pp. 439--447. ACM (2020)

\bibitem{21idreveal}
Cozzolino, D., R{\"{o}}ssler, A., Thies, J., Nie{\ss}ner, M., Verdoliva, L.:
  Id-reveal: Identity-aware deepfake video detection. In: International
  Conference on Computer Vision (ICCV). pp. 15088--15097. IEEE/CVF (2021)

\bibitem{20attention}
Dang, H., Liu, F., Stehouwer, J., Liu, X., Jain, A.K.: On the detection of
  digital face manipulation. In: Conference on Computer Vision and Pattern
  Recognition (CVPR). pp. 5780--5789. IEEE/CVF (2020)

\bibitem{deepfakes}
DeepFakes: \url{https://github.com/deepfakes/faceswap}

\bibitem{20retina}
Deng, J., Guo, J., Ververas, E., Kotsia, I., Zafeiriou, S.: Retinaface:
  Single-shot multi-level face localisation in the wild. In: Conference on
  Computer Vision and Pattern Recognition (CVPR). pp. 5202--5211. IEEE/CVF
  (2020)

\bibitem{20dfdc}
Dolhansky, B., Bitton, J., Pflaum, B., Lu, J., Howes, R., Wang, M.,
  Canton-Ferrer, C.: The deepfake detection challenge dataset. CoRR
  \textbf{abs/2006.07397} (2020)

\bibitem{19dfdcp}
Dolhansky, B., Howes, R., Pflaum, B., Baram, N., Canton-Ferrer, C.: The
  deepfake detection challenge (dfdc) preview dataset. CoRR
  \textbf{abs/1910.08854} (2019)

\bibitem{faceswap}
FaceSwap: \url{https://github.com/MarekKowalski/FaceSwap}

\bibitem{20frank}
Frank, J., Eisenhofer, T., Sch{\"{o}}nherr, L., Fischer, A., Kolossa, D., Holz,
  T.: Leveraging frequency analysis for deep fake image recognition. In:
  International Conference on Machine Learning (ICML). Proceedings of Machine
  Learning Research, vol.~119, pp. 3247--3258. PMLR (2020)

\bibitem{15rcnn}
Girshick, R.B.: Fast r-cnn. In: International Conference on Computer Vision
  (ICCV). pp. 1440--1448. IEEE (2015)

\bibitem{21inconsistency}
Gu, Z., Chen, Y., Yao, T., Ding, S., Li, J., Huang, F., Ma, L.: Spatiotemporal
  inconsistency learning for deepfake video detection. In: ACM International
  Conference on Multimedia (ACM MM). pp. 3473--3481. ACM (2021)

\bibitem{18rnn}
Guera, D., Delp, E.J.: Deepfake video detection using recurrent neural
  networks. In: International Conference on Advanced Video and Signal Based
  Surveillance (AVSS). pp.~1--6. IEEE (2018)

\bibitem{21lipforensics}
Haliassos, A., Vougioukas, K., Petridis, S., Pantic, M.: Lips don't lie: A
  generalisable and robust approach to face forgery detection. In: Conference
  on Computer Vision and Pattern Recognition (CVPR). pp. 5039--5049. IEEE/CVF
  (2021)

\bibitem{18resnet3d}
Hara, K., Kataoka, H., Satoh, Y.: Can spatiotemporal 3d cnns retrace the
  history of 2d cnns and imagenet? In: Conference on Computer Vision and
  Pattern Recognition (CVPR). pp. 6546--6555. IEEE/CVF (2018)

\bibitem{16resnet}
He, K., Zhang, X., Ren, S., Sun, J.: Deep residual learning for image
  recognition. In: Conference on Computer Vision and Pattern Recognition
  (CVPR). pp. 770--778. IEEE/CVF (2016)

\bibitem{18se}
Hu, J., Shen, L., Sun, G.: Squeeze-and-excitation networks. In: Conference on
  Computer Vision and Pattern Recognition (CVPR). pp. 7132--7141. IEEE/CVF
  (2018)

\bibitem{21dianet}
Hu, Z., Xie, H., Wang, Y., Li, J., Wang, Z., Zhang, Y.: Dynamic
  inconsistency-aware deepfake video detection. In: Proceedings of the
  International Joint Conference on Artificial Intelligence (IJCAI). pp.
  736--742. ijcai.org (2021)

\bibitem{20faceshifter}
Li, L., Bao, J., Yang, H., Chen, D., Wen, F.: Advancing high fidelity identity
  swapping for forgery detection. In: Conference on Computer Vision and Pattern
  Recognition (CVPR). pp. 5073--5082. IEEE/CVF (2020)

\bibitem{20facexray}
Li, L., Bao, J., Zhang, T., Yang, H., Chen, D., Wen, F., Guo, B.: Face x-ray
  for more general face forgery detection. In: Conference on Computer Vision
  and Pattern Recognition (CVPR). pp. 5000--5009. IEEE/CVF (2020)

\bibitem{18eyeblink}
Li, Y., Chang, M.C., Lyu, S.: In ictu oculi: Exposing ai created fake videos by
  detecting eye blinking. In: International Workshop on Information Forensics
  and Security (WIFS). pp.~1--7. IEEE (2018)

\bibitem{19fwa}
Li, Y., Lyu, S.: Exposing deepfake videos by detecting face warping artifacts.
  In: Conference on Computer Vision and Pattern Recognition Workshops (CVPRW).
  pp. 46--52. IEEE/CVF (2019)

\bibitem{20cdf}
Li, Y., Yang, X., Sun, P., Qi, H., Lyu, S.: Celeb-df: A large-scale challenging
  dataset for deepfake forensics. In: Conference on Computer Vision and Pattern
  Recognition (CVPR). pp. 3204--3213. IEEE/CVF (2020)

\bibitem{20cnn3d}
de~Lima, O., Franklin, S., Basu, S., Karwoski, B., George, A.: Deepfake
  detection using spatiotemporal convolutional networks. CoRR
  \textbf{abs/2006.14749} (2020)

\bibitem{21spsl}
Liu, H., Li, X., Zhou, W., Chen, Y., He, Y., Xue, H., Zhang, W., Yu, N.:
  Spatial-phase shallow learning: Rethinking face forgery detection in
  frequency domain. In: Conference on Computer Vision and Pattern Recognition
  (CVPR). pp. 772--781. IEEE/CVF (2021)

\bibitem{17cosine}
Loshchilov, I., Hutter, F.: Sgdr: Stochastic gradient descent with warm
  restarts. In: International Conference on Learning Representations (ICLR).
  OpenReview.net (2017)

\bibitem{19adamw}
Loshchilov, I., Hutter, F.: Decoupled weight decay regularization. In:
  International Conference on Learning Representations (ICLR). OpenReview.net
  (2019)

\bibitem{21hff}
Luo, Y., Zhang, Y., Yan, J., Liu, W.: Generalizing face forgery detection with
  high-frequency features. In: Conference on Computer Vision and Pattern
  Recognition (CVPR). pp. 16317--16326. IEEE/CVF (2021)

\bibitem{tsne}
Van~der Maaten, L., Hinton, G.: Visualizing data using t-sne. Journal of
  Machine Learning Research  \textbf{9}(11) (2008)

\bibitem{20twobranch}
Masi, I., Killekar, A., Mascarenhas, R.M., Gurudatt, S.P., AbdAlmageed, W.:
  Two-branch recurrent network for isolating deepfakes in videos. In: European
  Conference on Computer Vision (ECCV). Lecture Notes in Computer Science, vol.
  12352, pp. 667--684. Springer (2020)

\bibitem{19visualartifacts}
Matern, F., Riess, C., Stamminger, M.: Exploiting visual artifacts to expose
  deepfakes and face manipulations. In: Winter Applications of Computer Vision
  Workshops (WACVW). pp. 83--92. IEEE (2019)

\bibitem{20emotion}
Mittal, T., Bhattacharya, U., Chandra, R., Bera, A., Manocha, D.: Emotions
  don't lie: An audio-visual deepfake detection method using affective cues.
  In: ACM International Conference on Multimedia (ACM MM). pp. 2823--2832. ACM
  (2020)

\bibitem{19multitask}
Nguyen, H.H., Fang, F., Yamagishi, J., Echizen, I.: Multi-task learning for
  detecting and segmenting manipulated facial images and videos. In:
  International Conference on Biometrics Theory, Applications and Systems
  (BTAS). pp.~1--8. IEEE (2019)

\bibitem{19fsgan}
Nirkin, Y., Keller, Y., Hassner, T.: Fsgan: Subject agnostic face swapping and
  reenactment. In: International Conference on Computer Vision (ICCV). pp.
  7183--7192. IEEE/CVF (2019)

\bibitem{ntechlab}
NTech-Lab: \url{https://github.com/NTech-Lab/deepfake-detection-challenge}

\bibitem{20f3net}
Qian, Y., Yin, G., Sheng, L., Chen, Z., Shao, J.: Thinking in frequency: Face
  forgery detection by mining frequency-aware clues. In: European Conference on
  Computer Vision (ECCV). Lecture Notes in Computer Science, vol. 12357, pp.
  86--103. Springer (2020)

\bibitem{19ff}
R{\"{o}}ssler, A., Cozzolino, D., Verdoliva, L., Riess, C., Thies, J.,
  Nie{\ss}ner, M.: Faceforensics++: Learning to detect manipulated facial
  images. In: International Conference on Computer Vision (ICCV). pp. 1--11.
  IEEE/CVF (2019)

\bibitem{19cnngru}
Sabir, E., Cheng, J., Jaiswal, A., AbdAlmageed, W., Masi, I., Natarajan, P.:
  Recurrent convolutional strategies for face manipulation detection in videos.
  In: Conference on Computer Vision and Pattern Recognition Workshops (CVPRW).
  pp. 80--87. IEEE/CVF (2019)

\bibitem{17gradcam}
Selvaraju, R.R., Cogswell, M., Das, A., Vedantam, R., Parikh, D., Batra, D.:
  Grad-cam: Visual explanations from deep networks via gradient-based
  localization. In: International Conference on Computer Vision (ICCV). pp.
  618--626. IEEE (2017)

\bibitem{21dcl}
Sun, K., Yao, T., Chen, S., Ding, S., L, J., Ji, R.: Dual contrastive learning
  for general face forgery detection. CoRR  \textbf{abs/2112.13522} (2021)

\bibitem{19neuraltextures}
Thies, J., Zollh{\"{o}}fer, M., Nie{\ss}ner, M.: Deferred neural rendering:
  Image synthesis using neural textures. ACM Trans. Graph.  \textbf{38}(4),
  66:1--66:12 (2019)

\bibitem{16face2face}
Thies, J., Zollh{\"{o}}fer, M., Stamminger, M., Theobalt, C., Nie{\ss}ner, M.:
  Face2face: Real-time face capture and reenactment of rgb videos. In:
  Conference on Computer Vision and Pattern Recognition (CVPR). pp. 2387--2395.
  IEEE/CVF (2016)

\bibitem{20cnnaug}
Wang, S.Y., Wang, O., Zhang, R., Owens, A., Efros, A.A.: Cnn-generated images
  are surprisingly easy to spot... for now. In: Conference on Computer Vision
  and Pattern Recognition (CVPR). pp. 8692--8701. IEEE/CVF (2020)

\bibitem{19headpose}
Yang, X., Li, Y., Lyu, S.: Exposing deep fakes using inconsistent head poses.
  In: International Conference on Acoustics, Speech and Signal Processing
  (ICASSP). pp. 8261--8265. IEEE (2019)

\bibitem{18mixup}
Zhang, H., Ciss{\'{e}}, M., Dauphin, Y.N., Lopez-Paz, D.: Mixup: Beyond
  empirical risk minimization. In: International Conference on Learning
  Representations (ICLR). OpenReview.net (2018)

\bibitem{19zhang}
Zhang, X., Karaman, S., Chang, S.F.: Detecting and simulating artifacts in gan
  fake images. In: International Workshop on Information Forensics and Security
  (WIFS). pp.~1--6. IEEE (2019)

\bibitem{21multiattention}
Zhao, H., Zhou, W., Chen, D., Wei, T., Zhang, W., Yu, N.: Multi-attentional
  deepfake detection. In: Conference on Computer Vision and Pattern Recognition
  (CVPR). pp. 2185--2194. IEEE/CVF (2021)

\bibitem{21pcl}
Zhao, T., Xu, X., Xu, M., Ding, H., Xiong, Y., Xia, W.: Learning
  self-consistency for deepfake detection. In: International Conference on
  Computer Vision (ICCV). pp. 15003--15013. IEEE/CVF (2021)

\bibitem{17twostream}
Zhou, P., Han, X., Morariu, V.I., Davis, L.S.: Two-stream neural networks for
  tampered face detection. In: Conference on Computer Vision and Pattern
  Recognition Workshops (CVPRW). pp. 1831--1839. IEEE/CVF (2017)

\bibitem{21jointav}
Zhou, Y., Lim, S.N.: Joint audio-visual deepfake detection. In: International
  Conference on Computer Vision (ICCV). pp. 14780--14789. IEEE/CVF (2021)

\end{thebibliography}
